\documentclass[10pt,twocolumn,letterpaper]{article}

\usepackage[pagenumbers]{cvpr} 

\usepackage[dvipsnames]{xcolor}

\definecolor{cvprblue}{rgb}{0.21,0.49,0.74}
\usepackage[pagebackref,breaklinks,colorlinks,citecolor=cvprblue]{hyperref}
\usepackage{url}
\usepackage{graphicx}
\usepackage{amsmath}
\usepackage{amssymb}
\usepackage{booktabs}
\usepackage{url}
\usepackage{multirow}
\usepackage{color}
\usepackage{colortbl}

\usepackage[accsupp]{axessibility} 

\def\paperID{7153} 
\def\confName{CVPR}
\def\confYear{2024}

\title{DifFlow3D: Toward Robust Uncertainty-Aware Scene Flow Estimation with Diffusion Model}

\author
{Jiuming~Liu\textsuperscript{\rm 1}, Guangming~Wang\textsuperscript{\rm 2}, Weicai~Ye\textsuperscript{\rm 3}, Chaokang~Jiang\textsuperscript{\rm 4}, Jinru~Han\textsuperscript{\rm 1}, \\Zhe~Liu\textsuperscript{\rm 5}, Guofeng~Zhang\textsuperscript{\rm 3}, Dalong~Du\textsuperscript{\rm 4}, Hesheng~Wang\textsuperscript{\rm 1}\thanks{ Corresponding Authors.}\\
{\textsuperscript{\rm 1}Department of Automation, Shanghai Jiao Tong University}\qquad
{\textsuperscript{\rm 2}University of Cambridge} \\
{\textsuperscript{\rm 3}State Key Lab of CAD \& CG, Zhejiang University}\qquad
{\textsuperscript{\rm 4}PhiGent Robotics} \\
{\textsuperscript{\rm 5} MoE Key Lab of Artificial Intelligence, Shanghai Jiao Tong University}\\
\small{\texttt{\{liujiuming,wangguangming,liuzhesjtu,wanghesheng\}@sjtu.edu.cn}} \\
\small{\texttt{ts20060079a31@cumt.edu.cn}}\qquad \small{\texttt{maikeyeweicai@gmail.com }}\\
}

\begin{document}

\twocolumn[{%
    \renewcommand\twocolumn[1][]{#1}%
    \setlength{\tabcolsep}{0.0mm} 
    \maketitle
    \begin{center}
        \newcommand{\teaserwidth}{\textwidth}
    \vspace{-0.3in}
        \includegraphics[width=\linewidth]{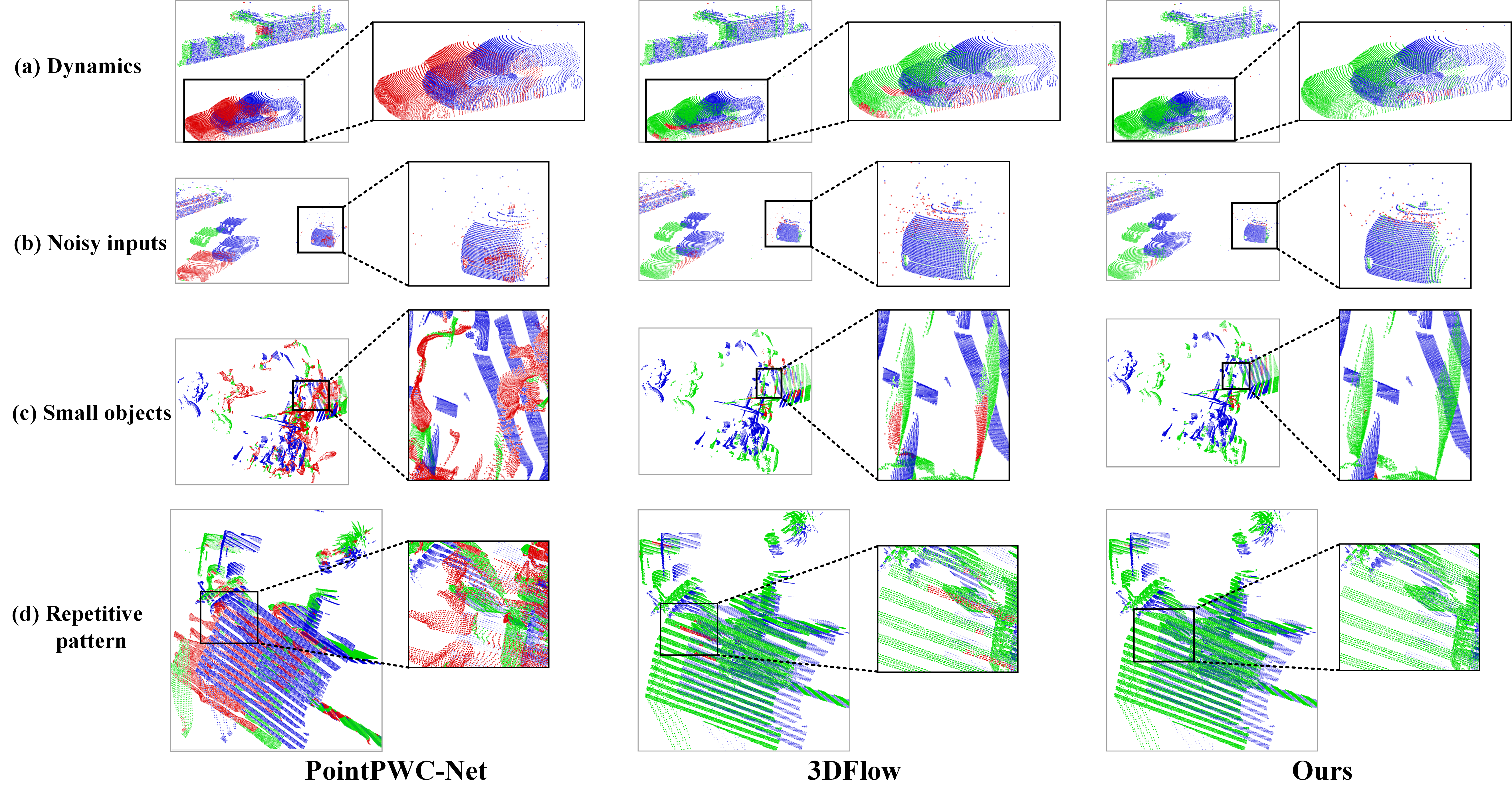}
      \vspace{-8mm}
        \captionof{figure}{\textbf{Comparison on challenging cases.} DifFlow3D predicts uncertainty-aware scene flow with diffusion model, which has stronger robustness for: \textbf{(a) dynamics, (b) noisy inputs, (c) small objects, and (d) repetitive patterns}. \textcolor{blue}{Blue}, \textcolor{green!50!gray}{green}, \textcolor{red}{red} points respectively indicate \textcolor{blue}{the first frame $PC_1$},  \textcolor{green!50!gray} {accurately estimated $PC_2$} ($PC_1$ warped by estimated flow), and \textcolor{red}{inaccurately estimated $PC_2$}. }
    \label{fig:vis}
    \end{center}
}]

\setcounter{footnote}{1}
\footnotetext{*Corresponding Author.}
\vspace{-6mm}
\begin{abstract}
\vspace{-3mm}
   Scene flow estimation, which aims to predict per-point 3D displacements of dynamic scenes, is a fundamental task in the computer vision field. However, previous works commonly suffer from unreliable correlation caused by locally constrained searching ranges, and struggle with accumulated inaccuracy arising from the coarse-to-fine structure. To alleviate these problems, we propose a novel uncertainty-aware scene flow estimation network (DifFlow3D) with the diffusion probabilistic model. Iterative diffusion-based refinement is designed to enhance the correlation robustness and resilience to challenging cases, e.g. dynamics, noisy inputs, repetitive patterns, etc. To restrain the generation diversity, three key flow-related features are leveraged as conditions in our diffusion model. Furthermore, we also develop an uncertainty estimation module within diffusion to evaluate the reliability of estimated scene flow. Our DifFlow3D achieves state-of-the-art performance, with $24.0\%$ and $29.1\%$ EPE3D reduction respectively on FlyingThings3D and KITTI 2015 datasets. Notably, our method achieves an unprecedented millimeter-level accuracy ($0.0078$m in EPE3D) on the KITTI dataset. Additionally, our diffusion-based refinement paradigm can be readily integrated as a plug-and-play module into existing scene flow networks, significantly increasing their estimation accuracy. Codes are released at \url{https://github.com/IRMVLab/DifFlow3D}.
\end{abstract}

\vspace{-8pt}
\section{Introduction}
\vspace{-2pt}
As a fundamental task in computer vision, scene flow refers to the 3D motion field estimated from consecutive images or point clouds. It provides the low-level perception of dynamic scenes and has various down-stream applications, such as autonomous driving \cite{najibi2022motion,gu2022rcp}, SLAM \cite{xie2024angular,zhu2023sni,liu2023translo}, and motion segmentation \cite{baur2021slim,deng2023long}. Early works focus on employing stereo \cite{huguet2007variational} or RGB-D images \cite{hadfield2011kinecting} as input. With the increasing application of 3D sensors, e.g. LiDAR, recent works commonly take point clouds directly as input.

As a pioneering work, FlowNet3D \cite{liu2019flownet3d} extracts hierarchical features with PointNet++ \cite{qi2017pointnet++}, and then regresses the scene flow iteratively. PointPWC \cite{wu2020pointpwc} further improves it with the Pyramid, Warping, and Cost volume structure \cite{sun2018pwc}. HALFlow \cite{wang2021hierarchical} follows them and introduces the attention mechanism for better flow embedding. However, these regression-based works commonly suffer from unreliable correlation \cite{liu2023regformer} and local optimum problems \cite{lu2021hregnet}. Reasons are mainly two folds: (1) \emph{K} Nearest Neighbor (KNN) is utilized for searching point correspondences in their networks, which can not take into account correct yet distant point pairs and also has matching noise \cite{fu2023pt}. (2) Another potential problem arises from the coarse-to-fine structure widely used in previous works \cite{liu2019flownet3d,wu2020pointpwc,wang2021hierarchical,wang2022matters}. Basically, an initial flow is estimated in the coarsest layer and then iteratively refined in higher resolutions. However, the performance of flow refinement highly relies on the reliability of initial coarse flow, since subsequent refinement is typically constrained within small spatial ranges around the initialization.

\begin{figure}[t]
\centering
\includegraphics[width=1.0\linewidth]{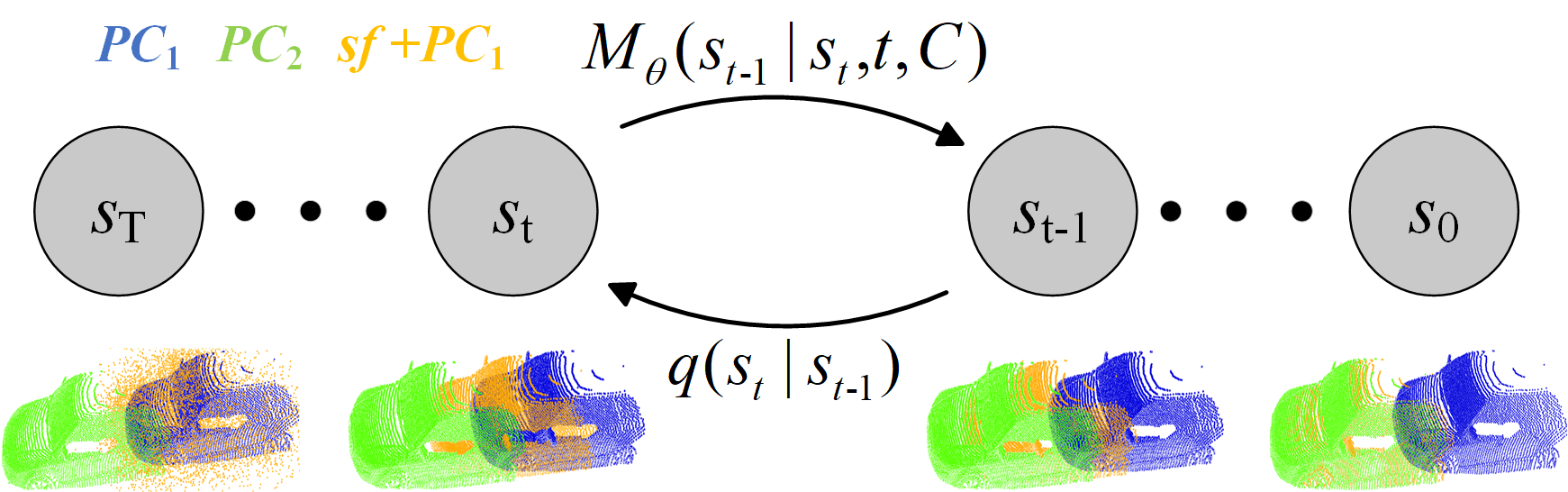}
\vspace{-5mm}
\caption{\textbf{An illustration of our diffusion for scene flow estimation.} During the forward process, we progressively add Gaussian noise on the ground truth flow residual ($s_{0}$). A neural network $M_{\theta}(\cdot, \cdot, \cdot)$ is trained to denoise the noisy flow residual $s_{t}$ at time $t$ based on condition information $C$. }
\vspace{-14pt}
\label{fig:diff}
\end{figure}

To address the unreliability issue, 3DFlow \cite{wang2022matters} designs an all-to-all point-gathering module with backward validation. Similarly, BPF \cite{cheng2022bi} and its extension MSBRN \cite{cheng2023multi} propose a bi-directional network with the forward-backward correlation. IHNet \cite{wang2023ihnet} leverages a recurrent network with high-resolution guidance and a re-sampling scheme. However, these networks mostly struggle with large computational costs due to their bi-directional association or recurrent iteration. In this paper, we discover that diffusion models can also enhance the correlation reliability and resilience to matching noise due to its denoising intrinsic (Fig. \ref{fig:vis}). Motivated by the findings in \cite{song2020score} that injecting random noise is helpful to jump out of local optimum, we re-formulate the deterministic flow regression task with a probabilistic diffusion model as illustrated in Fig. \ref{fig:diff}. Moreover, our method can serve as a plug-and-play module in previous scene flow networks, which is more generic and almost has no computational overhead (Section \ref{runtime conmparison}).

However, it is rather challenging to leverage a generative model in our task, due to the inherent generation diversity of diffusion models. Unlike the point cloud generation task requiring diverse output samples, scene flow prediction is a deterministic task that calculates precise per-point motion vectors. To tackle this problem, we leverage strong condition information to restrict the diversity and effectively control the generated flow. Specifically, a coarse sparse scene flow is first initialized, and then the flow residuals are iteratively generated with diffusion. In each diffusion-based refinement layer, coarse flow embedding, cost volume, and geometry encoding are all utilized as conditions. In this case, diffusion is applied to learn a probabilistic mapping from conditional inputs to the flow residual.

Furthermore, previous works rarely explore the confidence and reliability of the scene flow estimation. However, dense flow matching is prone to errors in the case of noise, dynamics, small objects, and repetitive patterns as in Fig. \ref{fig:vis}. Thus, it is significant to know whether each estimated point correspondence is trustworthy. Inspired by the recent success of uncertainty estimation in optical flow task \cite{truong2023pdc}, we propose a per-point uncertainty in the diffusion model to evaluate the reliability of our scene flow estimation.

Overall, the contributions of our paper are as follows:
\begin{itemize}
\item For robust scene flow estimation, we propose a novel plug-and-play diffusion-based refinement pipeline. To the best of our knowledge, this is the first work to leverage diffusion probabilistic model in the scene flow task.

\item We design strong conditional guidance by combining coarse flow embeddings, geometry encoding, and cross-frame cost volume, to control the generation diversity. 

\item To evaluate the reliability of our estimated flow and identify inaccurate point matching, we also introduce a per-point uncertainty estimation within our diffusion model. 

\item Our method outperforms all existing methods on both FlyingThings3D and KITTI datasets. Especially, our DifFlow3D achieves the millimeter-level End-point-error (EPE3D) on KITTI dataset for the first time. Compared with previous works, our method has stronger robustness for challenging cases, e.g., noisy inputs, dynamics, etc.
\end{itemize}


\section{Related Work}
\vspace{-2pt}
\textbf{Scene Flow Estimation.}  As the counterpart of optical flow in 3D vision, scene flow has witnessed remarkable advances in recent years. FlowNet3D \cite{liu2019flownet3d} is a pioneering work that firstly extracts point features with PointNet++ \cite{qi2017pointnet++}, and then regresses the per-point motion vector hierarchically. PointPWC \cite{wu2020pointpwc} further improves it with the Pyramid, Warping, and Cost volume structure \cite{sun2018pwc} in a patch-to-patch manner. FLOT \cite{puy2020flot} proposes an optimal transport module to establish the cross-frame association. PV-RAFT \cite{wei2021pv} fuses the point and voxel representations for both local and global feature extraction. To tackle the unreliable correlation, 3DFlow, Bi-PointFlowNet, and MSBRN \cite{wang2022matters,cheng2022bi,cheng2023multi} design the bi-directional correlation layer. SFGAN \cite{wang2022sfgan} proposes a scene flow network with the generative adversarial network (GAN). DELFlow \cite{peng2023delflow} proposes a dense efficient flow estimation network for large-scale points. PT-FlowNet \cite{fu2023pt} proposes a transformer-based pipeline to capture long-range dependencies. There are also some self-supervised methods \cite{baur2021slim, li2022rigidflow,jiang20223d,shen2023self,lang2023scoop,jiang20243dsflabelling}, without requiring ground truth flow annotations. In this paper, we rethink scene flow estimation with a different insight: \emph{How to find a more generic method improving the robustness and reliability of previous regression-based methods?} On the one hand, we introduce a denoising diffusion model with strong conditions. On the other hand, uncertainty estimation is proposed to evaluate the reliability of dense flow matching.

\textbf{Diffusion models in 3D vision.} Recently, diffusion model \cite{sohl2015deep,ho2020denoising} has gained significant attention, which gradually removes noise from a Gaussian distribution to the sample data distribution by learning from its reverse process. It has yielded great advancements in image generation \cite{saharia2022photorealistic,rombach2022high}, video synthesis \cite{esser2023structure,yu2023video}, and static point cloud generation and completion \cite{luo2021diffusion,vahdat2022lion}. DPM \cite{luo2021diffusion} first applies the diffusion model conditioned on a shape latent to point cloud generation. PVD \cite{zhou20213d} proposes a 3D shape generation and completion network through point-voxel diffusion. LION \cite{vahdat2022lion} combines VAE and two latent diffusion models (feature-based and point-based) to further enhance the generation abilities. \emph{However, it remains unclear whether diffusion can model dynamic point clouds, i.e. scene flow.}

\textbf{Uncertainty estimation.} Uncertainty estimation has been widely explored in stereo matching, Multi-View Stereo (MVS), and optical flow fields. UCS-Net \cite{cheng2020deep} proposes an uncertainty-aware cascaded network where uncertainty interval is inferred from depth probabilities to progressively sub-divide local depth ranges. SEDNet \cite{chen2023learning} designs a loss function for joint disparity and uncertainty estimation in deep stereo matching. ProbFlow \cite{wannenwetsch2017probflow} proposes to jointly estimate the optical flow and its uncertainty. Similarly, PDCNet \cite{truong2023pdc} enhances ProbFlow with a new training strategy in terms of self-supervised training. \emph{Nevertheless, the uncertainty of scene flow estimation has been rarely studied before.} In this paper, we design a per-point uncertainty to evaluate the reliability of our estimated flow.

\begin{figure*}
 \centering
 \includegraphics[width=1.0\linewidth]{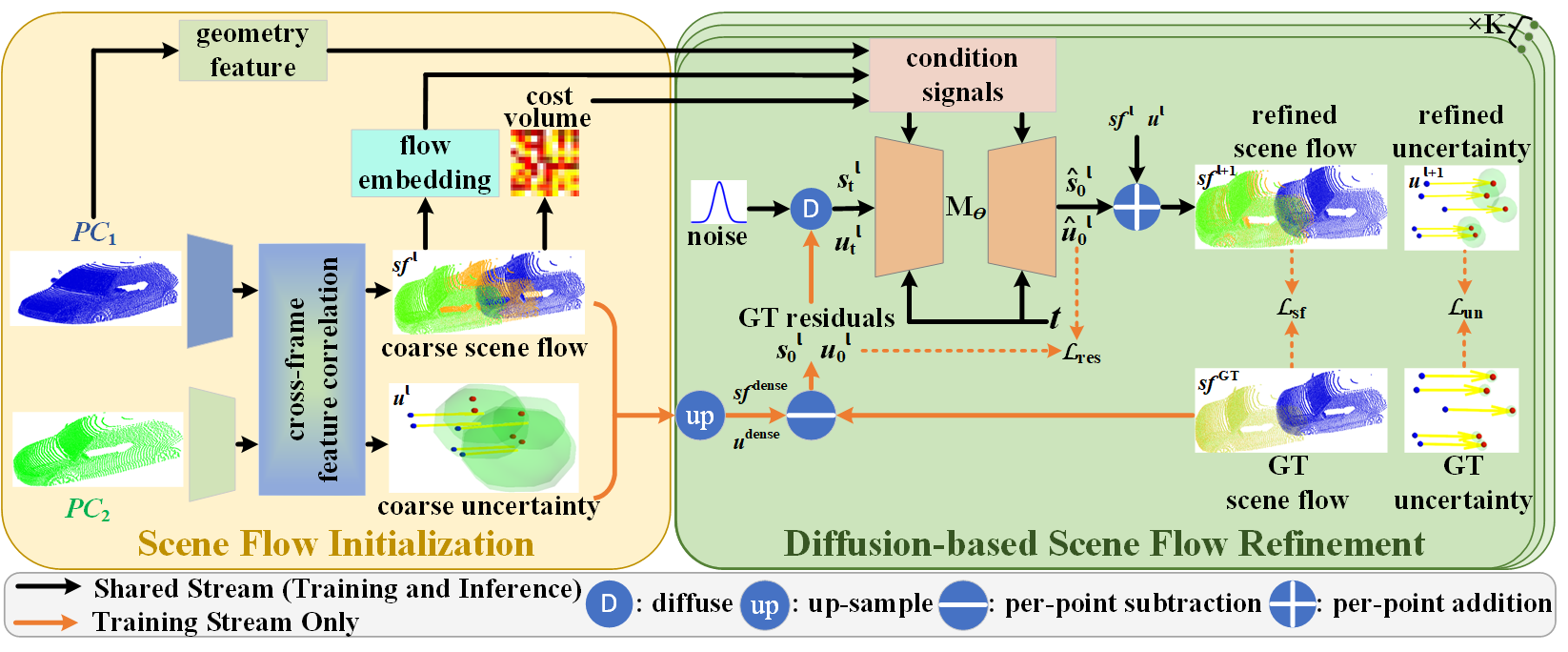}
 \vspace{-8mm}
 \caption{\textbf{The overall structure of DifFlow3D.} We first initialize a coarse sparse scene flow in the bottom layer. Then, iterative diffusion-based refinement layers with flow-related condition signals are applied to recover the denser flow residuals. A per-point uncertainty is also predicted jointly with scene flow to evaluate the reliability of our estimated flow.}
 \vspace{-14pt}
 \label{fig:pipeline}
\end{figure*}

\section{Method}
\vspace{-2pt}
Given two consecutive point cloud frames $PC_{1}\in\mathbb{R}^{N\times3}$ and $PC_{2}\in\mathbb{R}^{M\times3}$, scene flow $sf\in\mathbb{R}^{N\times3}$ indicates the 3D motion vector corresponding to each point in $PC_{1}$. 

As illustrated in Fig. \ref{fig:pipeline}, our network first hierarchically extracts point features and generates the initial scene flow in the coarsest layer (Section \ref{extraction}). Then, we utilize a diffusion model with strong condition signals to refine the coarse scene flow by predicting dense flow residuals from coarse to fine (Section \ref{refinement}). To evaluate the per-point prediction reliability of scene flow, uncertainty is also estimated jointly (Section \ref{uncertainty}) and propagated iteratively (Section \ref{archi}). Finally, the scene flow, flow residual, and uncertainty are all included for the network supervision (Section \ref{loss}). We will discuss each module specifically in the following sections.

\subsection{Feature Extraction and Flow Initialization}
\label{extraction}
\vspace{-2pt}
We adopt a hierarchical point feature extraction structure, where $PC_{1}$ and $PC_{2}$ are down-sampled through a series of set conv layers respectively. In each set conv layer, Farthest Point Sampling (FPS) is used to sample center points, and their neighboring point features are aggregated as:
\vspace{-2.5mm}
\begin{equation}
    \label{eq:feature gathering}
     {f_i^G} = \mathop{AGG}\limits_{m=1,2,\cdots,M} (MLP(({x_i^m}-{x_i})\oplus{f_i^m})),
     \vspace{-3mm}
\end{equation}
where $x_i$ is the $i$\mbox{-}th sampled center point. ${x_i^m}$ and ${f_i^m}$ represent its $m$\mbox{-}th neighbor point and feature respectively. ${f_i^G}$ means the output feature. $\oplus$ is the concatenation, and $AGG$ indicates the feature aggregation operation. It is worth noting that our method can be adapted to different initialization manners. The feature aggregation can be adopted by the max pooling in PointNet++ \cite{qi2017pointnet++} or Pointconv \cite{wu2019pointconv}.

After extracting point features hierarchically, we then leverage the cross-frame correlation \cite{wang2022matters, cheng2023multi} in the coarsest layer for the initialization of scene flow. However, the initialized scene flow $sf^{0}$ is sparse and inaccurate due to the low resolution and mismatching. Inspired by recent success and denoising properties of the diffusion model \cite{ho2020denoising}, we design a novel diffusion-based scene flow refinement module to recover the denser scene flow progressively. 

\subsection{Diffusion-based Scene Flow Refinement}
\label{refinement}
\vspace{-2pt}
It is non-trivial to directly recover scene flow from pure noise due to the significant discrepancies between their data distributions. Since coarse scene flow has been already obtained in the above flow initialization, what we exactly need is the flow residual between the initialized scene flow and the ground truth. Therefore, we formulate the scene flow residual as the diffusion latent variable, and multi-scale flow residuals are generated from the reverse process of diffusion model iteratively.

\textbf{Forward diffusion process.} While training, the forward diffusion process gradually adds Gaussian noise into the ground truth flow residual for $T$ timesteps via a Markov chain:
\vspace{-2mm}
\begin{equation}
    \label{eq:feature gathering}
     q(s_{1:T}^{i}|s_{0}^{i}) = \Pi_{t=1}^{T}  q(s_{t}^{i}|s_{t-1}^{i}),
     \vspace{-1mm}
\end{equation}
where $s_{0}^{i}$ indicates the ground truth flow residual for $i$\mbox{-}th point in $PC_{1}$. $s_{t}^{i}$ is the intermediate flow residual at timestamp $t$. Gaussian transition kernel $q(s_{t}^{i}|s_{t-1}^{i}) =\mathcal{N} (s_{t}^{i}; \sqrt{1-\beta_{t}} s_{t-1}^{i}, \beta_{t}I)$ is fixed by predefined hyper-parameters $\beta_{t}$. The forward process progressively adds noise to the ground truth flow residual $s_{0}^{i}$ and eventually turns it into a total Gaussian noise when $T$ is large enough.

In practice, coarse sparse flow is first up-sampled by the Three-Nearest Neighbors (Three-NN) as in \cite{wang2022matters, cheng2023multi}. The ground truth flow residual is then constructed through the subtraction between up-sampled coarse dense scene flow and the ground truth with the same resolution as in Fig. \ref{fig:pipeline}. 

\textbf{Reverse denoising process.} To generate the flow residual with robustness to outliers, we resort to the reverse process of diffusion model. Basically, the reverse process can be represented as a parameterized Markov chain starting from a random noise $p(s_{T}^{i})$:
\vspace{-2mm}
\begin{equation}
    \label{eq:feature gathering}
     p_{\theta}(s_{0:T}^{i}) = p(s_{T}^{i})\Pi_{t=1}^{T} p_{\theta}(s_{t-1}^{i}|s_{t}^{i}),
     \vspace{-2mm}
\end{equation}
where the reverse transition kernel $p_{\theta}(s_{t-1}^{i}|s_{t}^{i})$ is approximated with a neural network. Here, we follow \cite{tevet2022human} to directly learn the latent variable - flow residual $\hat{s}_{0}^{i}$ by the denoising network $M_{\theta}(s_{t-1}^{i}|s_{t}^{i},t)$.

However, there still remain challenges since scene flow estimation requires precise values and directions, rather than diverse outputs as in the point cloud generation task. Therefore, we constrain the generation diversity and control the reverse process by the powerful condition information $C$. In this way, the training objective of flow residual can be represented by:
\vspace{-3mm}
\begin{equation}
    \label{eq:feature gathering}
     \mathcal{L}_{res} := \mathbb{E}_{{s}_{0}^{i},t}[{\parallel {s}_{0}^{i}- M_{\theta}(s_{t}^{i},t,C) \parallel}_{2}^{2}],
     \vspace{-2mm}
\end{equation}

where $M_{\theta}(\cdot, \cdot, \cdot)$ is the denoising network.

\textbf{Design of condition signals.} 3DFlow \cite{wang2022matters} compares different design choices and investigates what really matters for 3D scene flow estimation. Inspired by their work, our diffusion condition signals include three key flow-related components: geometry feature $p^i$ extracted from $PC_1$, cross-frame cost volume $cv^i$, and coarse dense flow embeddings $e^i$. Cross-frame cost volume \cite{wang2021hierarchical} is constructed by warping the first point cloud $PC_1$ and then attentively correlating with $PC_2$. The augmented version \cite{cheng2023multi} with Gate Recurrent Unit (GRU) is used here for better cross-frame correlation. Coarse dense flow embeddings are generated through the set-upconv layer in \cite{liu2019flownet3d} from coarse sparse flow embeddings. Finally, the condition information is the concatenation of these features:
\vspace{-3mm}
\begin{equation}
    \label{eq:feature gathering}
    c^i= p^i\oplus cv^i \oplus e^i.
    \vspace{-2mm}
\end{equation}
Condition signals $C=\left\{c^i|c^i\in\mathbb{R}^{d}, i=1,...,n^{l}\right\}$ guide the reverse process to progressively produce the refined dense scene flow residual $\hat{s}_{0}^{i} = M_{\theta}(s_{t}^{i},t,C)$. The superscript $i$ in $s_{t}^{i}$ will be omitted below for the simplicity. $n^{l}$ is the point number of $PC_1$ at layer $l$.

\subsection{Uncertainty Estimation}
\label{uncertainty}
\vspace{-2pt}
Previous works rarely study the uncertainty and reliability of estimated scene flow. However, there exist point matching errors when associating two point cloud frames, especially for noisy inputs, dynamics, and repetitive patterns (Fig. \ref{fig:vis}). In these cases, estimated flow vectors from the network are not equally reliable in terms of different points. Therefore, it is crucial to make the network aware of whether each estimated point correspondence is trustworthy during the training. For this purpose, we propose a per-point uncertainty jointly estimated with scene flow in our diffusion model to evaluate the estimation reliability.

\begin{figure}[t]
\centering
\includegraphics[width=1.0\linewidth]{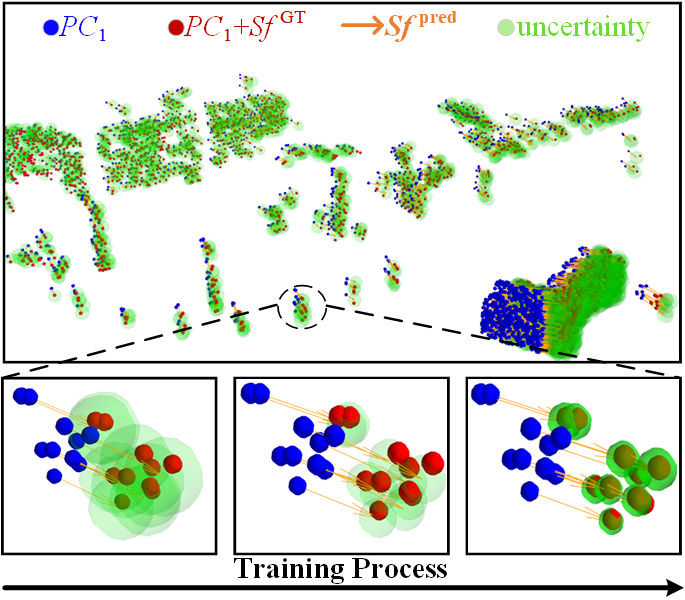}
\vspace{-6mm}
    \caption{\textbf{The visualization of uncertainty.} During the training process, our designed uncertainty intervals narrow progressively, which encourages predicted flow toward the ground truth.}
\vspace{-12pt}
\label{fig:un}
\end{figure}

\textbf{Estimation of uncertainty.} Cost volume represents the per-point matching degree between two point cloud frames. Therefore, in the coarsest layer, we output one additional channel as the initial uncertainty simultaneously with initial scene flow from cost volume as in Fig. \ref{fig:pipeline}. Then, in each refinement module, to control the flow residual generation more tightly, we also predict the uncertainty residuals using the same process as flow residuals through the denoising network of diffusion module:
\vspace{-8pt}
\begin{equation}
    \label{eq:feature gathering}
    \hat{s}_{0}^{l}, \hat{u}_{0}^{l}=M_{\theta}(s_{t}^{l},u_{t}^{l},t,C),
    \vspace{-8pt}
\end{equation}
where $\hat{u}_{0}^{l}$ represents the predicted uncertainty residual corresponding to each per-point flow residual $\hat{s}_{0}^{l}$. Finally, refined uncertainty in layer $l$ is then propagated into layer $l+1$ through the feature propagation process \cite{liu2019flownet3d}.

\textbf{Construction of the ground truth uncertainty.} The key issue is how to enable the network aware of uncertainty and how to supervise it toward the appropriate optimization direction. During the initial stages of training, the network has a relatively poor ability to estimate precise scene flow and tends to focus more on learning easier point correspondences. As the network converges, it has mastered the correspondence-finding prior to some extent for estimating more challenging cases. Therefore, our uncertainty intervals should also be narrowed dynamically throughout the training process. We obey this rule to construct the ground truth uncertainty as:
\vspace{-3mm}
\begin{equation}
    \label{eq:feature gathering}
    e_{ab}={\parallel sf^l - sf^{GT} \parallel}_2,
\end{equation}
\vspace{-6.6mm}
\begin{equation}
    \label{eq:feature gathering}
    e_{re}=|\frac{e_{ab}}{|sf^{GT}|}|,
\end{equation}
\vspace{-4mm}
\begin{equation}
{u}_{0}^{l}=
\begin{cases}
0,& \text{ $ e_{ab}<E_{1}, e_{re}<E_{2}$ } \\
1,& \text{ $ otherwise $ },
\end{cases}
\vspace{-1mm}
\end{equation}

where ${u}_{0}^{l}$ indicates the constructed uncertainty ground truth. $E_1$ and $E_2$ decay with the convergence of the network, which constrain the uncertainty intervals by absolute and relative estimation errors. As in Fig.\ref{fig:un}, uncertainty intervals represent the reliability range of each estimated flow and progressively encourage the estimated flow to optimize toward the ground truth. The supervision of uncertainty residual is similar to flow residual in the diffusion:
\vspace{-2mm}
\begin{equation}
    \label{eq:feature gathering}
     \mathcal{L}_{un} := \mathbb{E}_{{u}_{0}^{l},t}[{\parallel {u}_{0}^{l}- \hat{u}_{0}^{l} \parallel}_{2}^{2}].
     \vspace{-1mm}
\end{equation}
By incorporating uncertainty estimation into the training process, we enhance the network's ability to capture and quantify the uncertainty associated with each estimated scene flow, leading to more reliable and accurate results.

\subsection{Overall Architecture of DifFlow3D} 
\vspace{-2pt}
\label{archi}
We adopt the coarse-to-fine structure to generate the scene flow residuals and uncertainty residuals at different scales. The inputs of our diffusion-based refinement at layer $l$ are coarse sparse scene flow $sf^l$, coarse sparse uncertainty $u^l$ and condition signals $C^l$. Firstly, coarse sparse scene flow $sf^l$ is up-sampled to produce the coarse dense scene flow $sf^{dense}$. Similarly, coarse sparse uncertainty $u^l$ is up-sampled as coarse dense uncertainty $u^{dense}$. Dense flow residual $\hat{s}_{0}^{l}$, uncertainty residual $\hat{u}_{0}^{l}$ are then refined from the denoising network $M_{\theta}$. The refined dense scene flow (or uncertainty) at layer $l$ is generated by the per-point addition between coarse dense flow (or uncertainty) and its corresponding residual. Finally, the refined dense flow and uncertainty will be the inputs of layer $l+1$:
\vspace{-2mm}
\begin{equation}
    \label{eq:feature gathering}
    sf^{l+1}=sf^{dense}+\hat{s}_{0}^{l},
\end{equation}
\vspace{-6mm}
\begin{equation}
    \label{eq:feature gathering}
    u^{l+1}=u^{dense}+\hat{u}_{0}^{l}.
\end{equation}

\subsection{Training Objective}
\label{loss}
\vspace{-4pt}
We adopt a multi-scale supervision strategy in terms of scene flow, scene flow residuals, and uncertainty residuals. The scene flow loss is designed as:
\vspace{-2mm}
\begin{equation}
    \label{eq:feature gathering}
     \mathcal{L}_{sf}^{l} = {\parallel sf^{l}-sf^{GT}\parallel}_{2},
     \vspace{-2mm}
\end{equation}
where $sf^{l}$ and $sf^{GT}$ respectively indicate the estimated and ground truth scene flow at layer $l$. 

The overall loss function is the mixture of three parts:
\vspace{-2mm}
\begin{equation}
    \label{eq:feature gathering}
     \mathcal{L}^{l} = \lambda_{sf} \times \mathcal{L}_{sf}^{l}+\lambda_{res} \times\mathcal{L}_{res}^{l}+\lambda_{un} \times\mathcal{L}_{un}^{l},
\end{equation}
\vspace{-6mm}
\begin{equation}
    \label{eq:feature gathering}
     \mathcal{L} = \sum_{l=0}^{K} \alpha^{l} \times \mathcal{L}^{l},
     \vspace{-2mm}
\end{equation}
where $\alpha^{l}$ and $\mathcal{L}^{l}$ indicate the weight and total loss at layer $l$. $K$ is the number of our diffusion-based refinement layer.

\setlength{\tabcolsep}{0.2mm}
\begin{table*}[!t]
	\begin{center}
	
		\resizebox{1.00\textwidth}{!}
            {
		\begin{tabular}{lcccccc|cccccc}
		    \hline
			\toprule
			
			&\multicolumn{6}{c|}{FT3D$_{s}$ }&\multicolumn{6}{c}{KITTI$_{s}$ }\\
			 Method & EPE3D\textcolor{red}{$\downarrow$} & Acc3DS\textcolor{green!60!gray}{$\uparrow$} & Acc3DR\textcolor{green!60!gray}{$\uparrow$} & Outliers\textcolor{red}{$\downarrow$}& EPE2D\textcolor{red}{$\downarrow$}& Acc2D\textcolor{green!60!gray}{$\uparrow$}& EPE3D\textcolor{red}{$\downarrow$} & Acc3DS\textcolor{green!60!gray}{$\uparrow$}& Acc3DR\textcolor{green!60!gray}{$\uparrow$}& Outliers\textcolor{red}{$\downarrow$}& EPE2D\textcolor{red}{$\downarrow$}& Acc2D\textcolor{green!60!gray}{$\uparrow$}\\ \midrule

			FlowNet3D \cite{liu2019flownet3d}  & 0.1136 & 0.4125 & 0.7706& 0.6016& 5.9740 & 0.5692 & 0.1767& 0.3738 & 0.6677  & 0.5271  &  7.2141  & 0.5093  \\
			
			HPLFlowNet \cite{gu2019hplflownet}     & 0.0804 & 0.6144 & 0.8555& 0.4287& 4.6723 & 0.6764 & 0.1169 & 0.4783  & 0.7776  & 0.4103  &  4.8055  &  0.5938\\
			
			PointPWC-Net \cite{wu2020pointpwc}   & 0.0588   & 0.7379  & 0.9276 & 0.3424 & 3.2390 &  0.7994 & 0.0694 &  0.7281  &  0.8884 & 0.2648 &3.0062 &0.7673\\
			  
			 HALFlow \cite{wang2021hierarchical}    &  0.0492  &  0.7850&  0.9468 & 0.3083 &2.7555 &  0.8111 &  0.0622 & 0.7649 &0.9026 & 0.2492 & 2.5140 &  0.8128\\
			FLOT \cite{puy2020flot}    & 0.0520  &  0.7320 & 0.9270 &  0.3570 & ---&  --- & 0.0560 &  0.7550  &  0.9080  & 0.2420   & ---& ---\\
			
			 HCRF-Flow \cite{li2021hcrf}      & 0.0488 &0.8337  &  0.9507 &  0.2614  & 2.5652& 0.8704 & 0.0531& 0.8631&   0.9444 &0.1797  &  2.0700& 0.8656  \\
			
			PV-RAFT  \cite{wei2021pv}  & 0.0461 & 0.8169  & 0.9574  &  0.2924 & ---& ---  & 0.0560 & 0.8226  & 0.9372  &  0.2163 & ---& ---    \\

			FlowStep3D  \cite{kittenplon2021flowstep3d}  & 0.0455 & 0.8162  & 0.9614  &  0.2165 & ---& ---  & 0.0546 &0.8051  & 0.9254  & 0.1492  &  ---&     ---  \\
            
            RCP \cite{gu2022rcp} & 0.0403 & 0.8567 & 0.9635   &   0.1976   &  --- &   --- & 0.0481 & 0.8491 & 0.9448   &   0.1228   &  --- &   ---
            \\
			
			 3DFlow \cite{wang2022matters}    & 0.0281 & 0.9290 & 0.9817   &   0.1458   &  1.5229 &   0.9279 & 0.0309 &  0.9047  &0.9580    &    0.1612   &    1.1285 &       0.9451
            \\ 
            
             BPF \cite{cheng2022bi}    & 0.0280 & 0.9180 & 0.9780   &   0.1430   &  1.5820 &   0.9290 & 0.0300 & 0.9200 & 0.9600   &   0.1410   &  1.0560 &   0.9490
            \\
            
             PT-FlowNet \cite{fu2023pt}    & 0.0304 & 0.9142 & 0.9814   &   0.1735   &  1.6150 &   0.9312 & 0.0224 & 0.9551 & 0.9838   &   0.1186   &  0.9893 &   0.9667
            \\
             IHNet \cite{wang2023ihnet}    & 0.0191 & 0.9601 & 0.9865   &   0.0715   &  1.0918 &   0.9563 & 0.0122 & 0.9779 & 0.9892   &   0.0913   &  0.4993 &   0.9862
            \\

             MSBRN \cite{cheng2023multi}    & 0.0150 & 0.9730 & 0.9920   &   0.0560   &  0.8330 &   0.9700 & 0.0110 & 0.9710 & 0.9890   &   0.0850   &  0.4430 &   0.9850 
            \\
   
            Ours  
				&\bf0.0114 & \bf 0.9836
				&\bf 0.9949 & \bf 0.0350	
				& \bf 0.6220 & \bf 0.9824
				& \bf0.0078& \bf0.9817
				&\bf 0.9924 &  \bf0.0795
				&\bf 0.2987 & \bf0.9932\\
            
            \bottomrule
            \hline
			
		\end{tabular}
		}
	\vspace{-8pt}
	\caption{\textbf{Comparison results on FlyingThings3D and KITTI datasets without occlusion \cite{gu2019hplflownet}.} The best results are in bold. Our method has $\bf{24.0}\%$ and $\bf{29.1\%}$ EPE3D reduction respectively compared with previous works.}
	\label{table:flyingthing3d}
	\end{center}
	\vspace{-6mm}
	
\end{table*}

\section{Experiment}
\vspace{-2pt}
\subsection{Datasets and Implementation Details}
\vspace{-2pt}
We follow previous works \cite{liu2019flownet3d,gu2019hplflownet,puy2020flot,wei2021pv,wang2022matters,cheng2022bi} to train our DifFlow3D on synthetic FlyingThings3D dataset \cite{mayer2016large}, and evaluate the model on both FlyingThings3D and KITTI \cite{menze2015object} datasets. All the experiments are conducted on a single RTX 3090 GPU. The Adam optimizer is adopted with $\beta_1 = 0.9, \beta_2 = 0.999$. The learning rate is initially set as 0.0001 and reduced by 0.8 every 10 epochs. We leverage DDIM \cite{song2020denoising} with the total timestamp $T$ = 1000. Refinement layer number $K$ is 3. Gaussian noise is set as $\sigma$ = 1. We imitate the learning rate step decay strategy for $E_{1}$ and $E_{2}$, which are initially set as 0.5, and decay with the rate 0.8 until 0.02. Our evaluation metrics include EPE3D (m), Acc3DS, Acc3DR, Outliers, EPE2D (px), and Acc2D, commonly used in previous works \cite{liu2019flownet3d,wang2022matters,cheng2022bi}. 

\setlength{\tabcolsep}{0.2mm}
\begin{table}[t]
	\begin{center}
		
		
		\resizebox{1.00\columnwidth}{!}
            {
		\begin{tabular}{clcccccc}
		    \hline
			\toprule
			
		Dataset& Method  & Sup. & EPE3D\textcolor{red}{$\downarrow$} & Acc3DS\textcolor{green!60!gray}{$\uparrow$}& Acc3DR\textcolor{green!60!gray}{$\uparrow$}& Outliers\textcolor{red}{$\downarrow$}\\ \midrule

			& FlowNet3D \cite{liu2019flownet3d}  &Full  & 0.169 & 0.254 & 0.579 & 0.789\\

			&  FLOT \cite{puy2020flot}     &Full & 0.156  &  0.343 & 0.643  & 0.700 \\
			
		& OGSFNet \cite{ouyang2021occlusion}  &Full & 0.163 & 0.551 & 0.776 & 0.518\\
			
			&  FESTA \cite{wang2021festa}    &Full & 0.111  &  0.431 & 0.744  & --- \\
			
		    \multirow{-2}{*}{\begin{tabular}[c]{@{}c@{}}FT3D$_{o}$ \end{tabular}}

			&  3DFlow  \cite{wang2022matters}   &Full & 0.063 & 0.791 & 0.909  & 0.279 \\ 

            &  BPF \cite{cheng2022bi}  &Full & 0.073 & 0.791 & 0.896  & 0.274 \\
            &  MSBRN \cite{cheng2023multi}   &Full & 0.053 & 0.836 & 0.926  & 0.231  \\

            & Ours    &Full & \bf0.043 &\bf0.891  & \bf0.944 & \bf0.133 \\ 
			\midrule
			
			& FlowNet3D \cite{liu2019flownet3d}  &Full & 0.173 & 0.276 & 0.609  & 0.649\\
			
			&  FLOT \cite{puy2020flot}     &Full & 0.110  &  0.419 & 0.721   & 0.486\\
		
		& OGSFNet \cite{ouyang2021occlusion}  &Full & 0.075 & 0.706 & 0.869 & 0.327\\
			
			&  FESTA \cite{wang2021festa}     &Full & 0.097  &  0.449 & 0.833  & --- \\

			& 3DFlow \cite{wang2022matters}     &Full  &0.073  &0.819  & 0.890 & 0.261  \\ 

            \multirow{-3}{*}{\begin{tabular}[c]{@{}c@{}}KITTI$_{o}$\end{tabular}}
            &  BPF  \cite{cheng2022bi}  &Full & 0.065 & 0.769 & 0.906  & 0.264 \\
            
            &  SCOOP  \cite{lang2023scoop}  &Self & 0.063 & 0.797 & 0.910  & 0.244 \\
            
        &  MSBRN \cite{cheng2023multi}   &Full & 0.044 & 0.873 & 0.950  & 0.208  \\
   
            & Ours   &Full  &\bf0.031  &\bf0.955  & \bf0.966 & \bf0.108  \\ 
			\bottomrule
			\hline
		\end{tabular}
		}
	\vspace{-8pt}
	\caption{\textbf{Comparison results on FlyingThings3D and KITTI datasets with occlusion \cite{liu2019flownet3d}.} ``Self" and ``Full" respectively mean self-supervised and fully-supervised training. Our method outperforms previous works by \textbf{$\bf{18.9}\%$} and \textbf{$\bf{29.5}\%$} in terms of EPE3D.}
	\label{table:flyingthing3d_by_flownet3d}
	\end{center}
	\vspace{-6mm}
\end{table}

\subsection{Comparison with State-of-the-Art Methods}
\vspace{-2pt}
To comprehensively demonstrate the superiority of our method, we show comparison results with two pre-process settings: without occlusion \cite{gu2019hplflownet} and with occlusion \cite{liu2019flownet3d}. 

\textbf{Comparison on point clouds without occlusion.} We compare our DifFlow3D with a series of state-of-the-art (SOTA) methods on both FlyingThings3D \cite{mayer2016large} and KITTI scene flow \cite{menze2015object} datasets without occlusion. Table \ref{table:flyingthing3d} demonstrates that our DifFlow3D outperforms all previous methods in terms of both 3D and 2D metrics on two datasets. Compared with the recent SOTA method MSBRN \cite{cheng2023multi}, our model reduces the End-point-error (EPE3D) by 24.0\% and 29.1\% respectively on FlyingThings3D and KITTI datasets. It is worth noting that DifFlow3D also has the best generalization capability. Only trained on the Flyingthings3D synthetic dataset, our method firstly achieves millimeter-level (0.0078m) EPE3D metric on the KITTI dataset.

\textbf{Comparison on point clouds with occlusion.} We also evaluate our model on the datasets with occlusion prepared by \cite{liu2019flownet3d}. Experiment results in Table \ref{table:flyingthing3d_by_flownet3d} show that our DifFlow3D still surpasses all previous methods on both two datasets. Notably, our evaluation results on the KITTI dataset outperform previous methods by a large margin. Only trained on FT3D$_{O}$, our method reduces EPE3D by $29.5\%$ and outliers by $48.1\%$ on KITTI. This demonstrates the excellent generalization capability of our method.


\begin{figure*}[t]
\centering
\includegraphics[width=1.0\linewidth]{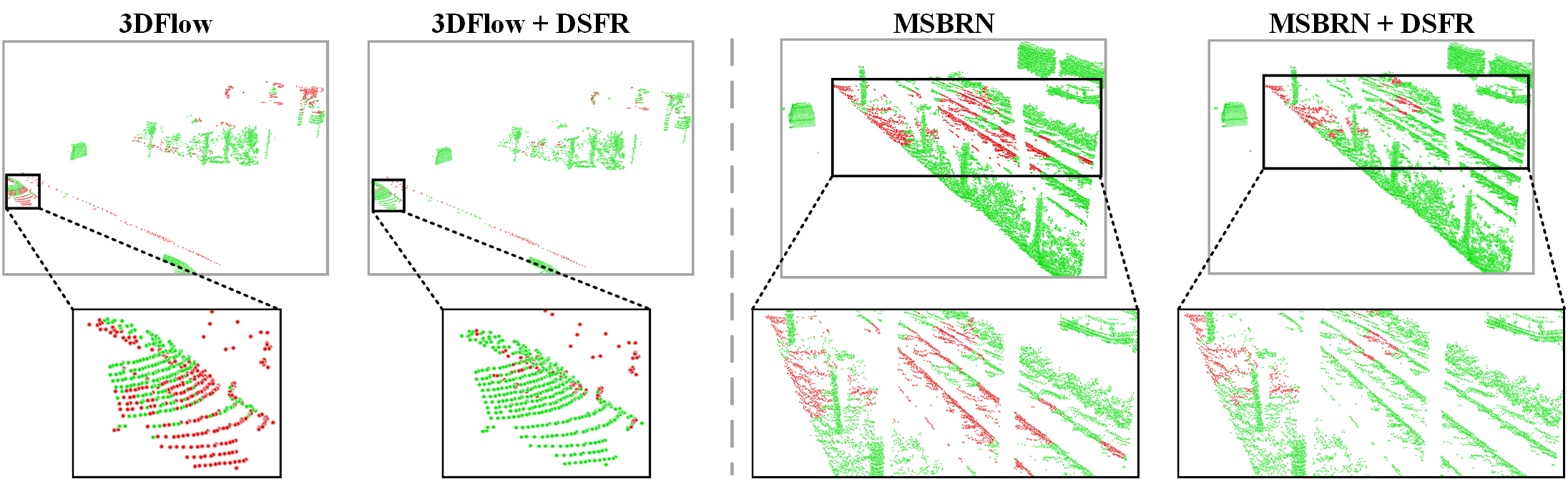}
\vspace{-8pt}
\caption{\textbf{Visualization results w/o or with our Diffusion-based Scene Flow Refinement (DSFR).} For better comparison, we only visualize the estimated $PC_2$ by warping $PC_1$ with estimated scene flow. \textcolor{green!50!gray}{green}, \textcolor{red}{red} points respectively indicate \textcolor{green!50!gray} {accurately estimated $PC_2$} and \textcolor{red}{inaccurately estimated $PC_2$ } (measured by Acc3DR).}
\vspace{-3pt}
\label{fig:visual1}
\end{figure*}

\setlength{\tabcolsep}{0.05mm}
\begin{table*}[!t]
	\begin{center}
		
		\resizebox{1.00\textwidth}{!}
            {
		\begin{tabular}{lclccccc|lccccc}
		    \hline
			\toprule
			
			 &&\multicolumn{6}{c|}{\begin{tabular}[c]{@{}c@{}}FT3D$_{s}$\end{tabular}}&\multicolumn{6}{c}{\begin{tabular}[c]{@{}c@{}}KITTI$_{s}$\end{tabular}}\\
			  \multirow{-2}{*}{\begin{tabular}[c]{@{}c@{}}Method \end{tabular}} &\multirow{-2}{*}{\begin{tabular}[c]{@{}c@{}}Training \end{tabular}} & EPE3D\textcolor{red}{$\downarrow$} & Acc3DS\textcolor{green!60!gray}{$\uparrow$}& Acc3DR\textcolor{green!60!gray}{$\uparrow$}& Outliers\textcolor{red}{$\downarrow$}& EPE2D\textcolor{red}{$\downarrow$}& Acc2D\textcolor{green!60!gray}{$\uparrow$}& EPE3D\textcolor{red}{$\downarrow$} & Acc3DS\textcolor{green!60!gray}{$\uparrow$}& Acc3DR\textcolor{green!60!gray}{$\uparrow$}& Outliers\textcolor{red}{$\downarrow$}& EPE2D\textcolor{red}{$\downarrow$}& Acc2D\textcolor{green!60!gray}{$\uparrow$}\\ \midrule
			
		    3DFlow \cite{wang2022matters}  & Quarter    & 0.0317   &  0.9109&0.9757   & 0.1673 & 1.7436& 0.9108   &0.0332  & 0.8931  & 0.9528  &  0.1690  & \bf1.2186& 0.9373\\ 
		    
		      \cellcolor{gray!20}3DFlow+DSFR  & \cellcolor{gray!20}Quarter
				&\cellcolor{gray!20}{\bf0.0297 \textcolor{green!50!gray}{($\downarrow$  6.3\%)}}  & \cellcolor{gray!20}\bf 0.9207
				&\cellcolor{gray!20}\bf 0.9785 & \cellcolor{gray!20}\bf 0.1548	
				& \cellcolor{gray!20}\bf 1.6344 & \cellcolor{gray!20}\bf 0.9188 &\cellcolor{gray!20}{\bf0.0316 \textcolor{green!50!gray}{($\downarrow$ 4.8\%)}} & \cellcolor{gray!20}\bf 0.9028
				&\cellcolor{gray!20}\bf 0.9634 & \cellcolor{gray!20}\bf 0.1604	
				&  \cellcolor{gray!20}1.2247 & \cellcolor{gray!20} \bf0.9413\\

			3DFlow \cite{wang2022matters}   & Complete  & 0.0281 & 0.9290 & 0.9817   &   0.1458   &  1.5229 &   0.9279 & 0.0309 &  0.9047  &0.9580    &    0.1612   &    1.1285 &       0.9451
            \\ 
             \cellcolor{gray!20}3DFlow+DSFR  &\cellcolor{gray!20} Complete
				&\cellcolor{gray!20}{\bf0.0242 \textcolor{green!50!gray}{($\downarrow$13.9\%)}} &\cellcolor{gray!20}\bf0.9494 
				&\cellcolor{gray!20}\bf0.9860&\cellcolor{gray!20}\bf0.1166
				&\cellcolor{gray!20}\bf1.3201 & \cellcolor{gray!20}\bf0.9459& \cellcolor{gray!20}{\bf0.0251  \textcolor{green!50!gray}{($\downarrow$18.8\%)}}& \cellcolor{gray!20}\bf0.9398
				&\cellcolor{gray!20}\bf 0.9793 &  \cellcolor{gray!20}\bf0.1302
				&\cellcolor{gray!20}\bf 0.9761 & \cellcolor{gray!20}\bf0.9686\\

            BPF \cite{cheng2022bi}  & Complete   & 0.0280 & 0.9180 & 0.9780   &   0.1430   &  1.5820 &   0.9290& 0.0300 & 0.9200 & 0.9600   &   0.1410   &  1.0560 &   0.9490 
            \\
            \cellcolor{gray!20}BPF+DSFR &\cellcolor{gray!20} Complete     & \cellcolor{gray!20}{\bf0.0247 \textcolor{green!50!gray}{($\downarrow$11.8\%)}} & \cellcolor{gray!20}\bf0.9390   &   \cellcolor{gray!20}\bf0.9840   &  \cellcolor{gray!20}\bf0.1192 &  \cellcolor{gray!20}\bf1.3749&
             \cellcolor{gray!20}\bf0.9468& \cellcolor{gray!20}{\bf0.0265 \textcolor{green!50!gray}{($\downarrow$11.7\%)}} & \cellcolor{gray!20}\bf0.9355& 
             \cellcolor{gray!20}\bf0.9672  &  
             \cellcolor{gray!20}\bf0.1290   & 
             \cellcolor{gray!20}\bf1.0527 & 
             \cellcolor{gray!20}\bf0.9637
            \\
            
              MSBRN \cite{cheng2023multi}  & Complete   & 0.0150 & 0.9730 & 0.9920   &   0.0560   &  0.8330 &   0.9700 & 0.0110 & 0.9710 & 0.9890   &   0.0850   &  0.4430 &   0.9850\\
   
             \cellcolor{gray!20}MSBRN+DSFR & \cellcolor{gray!20}Complete  &\cellcolor{gray!20}{\bf0.0114 \textcolor{green!50!gray}{($\downarrow$ 24.0\%)}} & \cellcolor{gray!20}\bf 0.9836&\cellcolor{gray!20}\bf 0.9949 & \cellcolor{gray!20}\bf 0.0350	& \cellcolor{gray!20}\bf 0.6220 & \cellcolor{gray!20}\bf 0.9824&\cellcolor{gray!20}{\bf0.0078 \textcolor{green!50!gray}{($\downarrow$29.1\%)}}& \cellcolor{gray!20}\bf0.9817&\cellcolor{gray!20}\bf 0.9924 & \cellcolor{gray!20} \bf0.0795&\cellcolor{gray!20}\bf 0.2987 & \cellcolor{gray!20}\bf0.9932\\

			\bottomrule
			\hline
		\end{tabular}
		}
	\vspace{-6pt}
	\caption{\textbf{The plug-and-play capability of our methods.} Our Diffusion-based Scene Flow Refinement (DSFR) can effectively improve the accuracy introduced into recent methods on both FlyingThings3D and KITTI datasets. The best results are in bold. }
		
		\label{table:general}
	\end{center}
	\vspace{-8mm}
\end{table*}

\subsection{Plug-and-Play on Previous Works}
\vspace{-2pt}
It is extremely encouraging that our diffusion-based refinement can serve as a plug-and-play module, improving the estimation accuracy of several recent SOTA works \cite{wang2022matters,cheng2022bi,cheng2023multi}. In Table \ref{table:general}, we replace their refinement modules with our diffusion-based one. 3DFlow \cite{wang2022matters} first trains their model on a quarter of the data, then finetunes results on the complete dataset. For the comprehensive comparison, we also show experiment results on both quarter and complete training sets. As shown in Table \ref{table:general}, our method respectively has 6.3\% and 13.9\% EPE3D reduction on the quarter and complete training set of FT3D$_{s}$ dataset. Similarly, introducing our diffusion refinement strategy into the GRU in BPF \cite{cheng2022bi} and MSBRN \cite{cheng2023multi} can also significantly improve their accuracy. Also, they innovate the cross-frame association with specific designs. Instead, our diffusion-based refinement is more generic in terms of different network designs or challenging cases. As in Fig. \ref{fig:visual1}, estimations for dynamic cars or bushes with complex patterns are more accurate by adding our Diffusion-based Scene Flow Refinement (DSFR).


\subsection{Ablation Study}
\vspace{-2pt}
Extensive experiments are conducted to validate the effectiveness of each proposed component. 

\textbf{Designs of diffusion model.} We first remove the diffusion refinement module, remaining the coarsest initialization stage. As in Table \ref{table:ablation} (a), there is a $0.14$m EPE3D increase. We also compare different denoising network choices (MLP \cite{wang2022matters}, point transformer \cite{nichol2022point}, or GRU \cite{cheng2023multi}). GRU has the best performance, partly because the recurrent network is more suitable for iterative refinement.

\textbf{Uncertainty.} As illustrated in Table \ref{table:ablation} (b), without our proposed uncertainty, there is $28.9\%$ higher EPE3D. Our uncertainty ground truth is designed as a binary value. We also conduct the experiment by substituting it with a continuous value ranging from 0 to 1. However, worse estimation results can be seen. To better guide the flow regression, we jointly estimate uncertainty and scene flow as in Section \ref{uncertainty}. If we predict uncertainty and scene flow by two separate fully connected layers from cost volume, there is a significant EPE3D increase. This is because uncertainty can not tightly control the flow due to the indirect connection and even mislead the flow learning in the wrong direction.

\textbf{Condition signals.} Conditional information is extremely significant in guiding the flow residual refinement module. We remove each of our designed condition signals to validate their importance as in Table \ref{table:ablation} (c). Without cost volume as guidance, our model has obviously worse estimation (4.5 times larger EPE3D), since cost volume precisely represents the per-point correlation between two frames. Coarse flow embedding can also lead the flow generation because the coarse matching in the low-resolution layer has a more global understanding of cross-frame correlation. Thus, if we remove it, there is a significant accuracy drop. The geometry feature has low-level 3D structural information in dynamic scenes. $31.6\%$ higher EPE3D can be seen without the geometry feature as guidance.

\setlength{\tabcolsep}{0.5mm}
\begin{table}[!t]
	\footnotesize
	\begin{center}
		\resizebox{1.0\columnwidth}{!}
		{
	\begin{tabular}{l||cccc}
	            \hline
				\toprule
				Method& EPE3D\textcolor{red}{$\downarrow$} & Acc3DS\textcolor{green}{$\uparrow$}& Acc3DR\textcolor{green}{$\uparrow$}& Outliers\textcolor{red}{$\downarrow$}\\
				\noalign{\smallskip}
				\hline\hline
				\noalign{\smallskip}
			
				\bf{(a) Diffusion}\\
				
				w/o diffusion refinement    
				&0.1559 &0.0777
				&0.4462 & 0.8637\\

				with transformer-based $M_{\theta}$ \cite{nichol2022point}     
				&0.0374 & 0.8465
				&0.9638 & 0.1862\\
				
				with MLP-based $M_{\theta}$ \cite{wang2022matters}    
				&0.0244 &0.9492
				&0.9858 & 0.1172\\

				Ours (with GRU-based $M_{\theta}$ \cite{cheng2023multi})     
				 &\bf0.0114 & \bf 0.9836
				&\bf 0.9949 & \bf 0.0350
				\\
				\cline{1-5}\noalign{\smallskip}
				
				\bf{(b) Uncertainty} \\
				
				w/o uncertainty 
				&0.0147 & 0.9753
				&0.9929 & 0.0515

				\\
				with continuous uncertainty GT   
				& 0.0178& 0.9669
				&0.9906 & 0.0774\\
				
				with separately estimated uncertainty     
				&0.0170 & 0.9692
				&0.9914 & 0.0661\\
				
				Ours (full)     
				 &\bf0.0114 & \bf 0.9836
				&\bf 0.9949 & \bf 0.0350\\
				\cline{1-5}\noalign{\smallskip}
				\bf{(c) Condition}\\
				
				w/o cost volume    
				&0.0621 & 0.6821
				&0.9098 & 0.4038\\
				
				w/o coarse flow embedding
				&0.0245 & 0.9415 
				&0.9817 & 0.1058\\
				
				w/o geometry feature
				&0.0150 & 0.9723
				&0.9921 & 0.0603\\
				
				Ours (full, with all three conditions)     
				 &\bf0.0114 & \bf 0.9836
				&\bf 0.9949 & \bf 0.0350\\
				\bottomrule
				\hline
            	
			\end{tabular}
		}
	
	\vspace{-8pt}
	\caption{\textbf{Ablation studies on FlyingThings3D prepared by \cite{gu2019hplflownet}.}}
	\vspace{-3mm}
	\label{table:ablation}
	\end{center}
\end{table}

\setlength{\tabcolsep}{3mm}
\begin{table}[!t]
	\begin{center}
		\resizebox{1.00\columnwidth}{!}
            {
		\begin{tabular}{l|c|l|c}
		    \hline
			\toprule
			Method& Runtime  & Method & Runtime \\ \midrule
		
			FLOT \cite{cheng2023multi} &289.6ms & PV-RAFT \cite{wei2021pv} & 781.1ms \\
			FlowStep3D \cite{kittenplon2021flowstep3d} & 972.7ms &RCP \cite{gu2022rcp} & 2854.6ms \\
			 MSBRN \cite{cheng2023multi} &221.1ms& Ours & 228.3ms\\
		    
			\bottomrule
			\hline
		\end{tabular}
		}
	\vspace{-8pt}
	\caption{\textbf{Runtime comparison.} Compared with our baseline MSBRN \cite{cheng2023multi}, our proposed diffusion-based refinement only brings minimal computational overhead.}
	\label{table:time}
	\end{center}
	\vspace{-3mm}
\end{table}

\subsection{Runtime Comparison}
\label{runtime conmparison}
\vspace{-2pt}
We compare our DifFlow3D with previous methods on efficiency. All results are evaluated on a single RTX 3090 GPU. Table \ref{table:time} shows that DifFlow3D has highly competitive efficiency. Notice that our baseline here is based on MSBRN \cite{cheng2023multi}, where we combine its flow initialization with our diffusion-based refinement. Compared with MSBRN, our method has only a slightly more computational overhead ($3.3\%$), but much higher estimation accuracy ($24.0\%$ and $29.1\%$ on FT3D$_{s}$ and KITTI$_{s}$ as in Table \ref{table:flyingthing3d}).

\begin{figure}[t]
\centering
\includegraphics[width=1.0\linewidth]{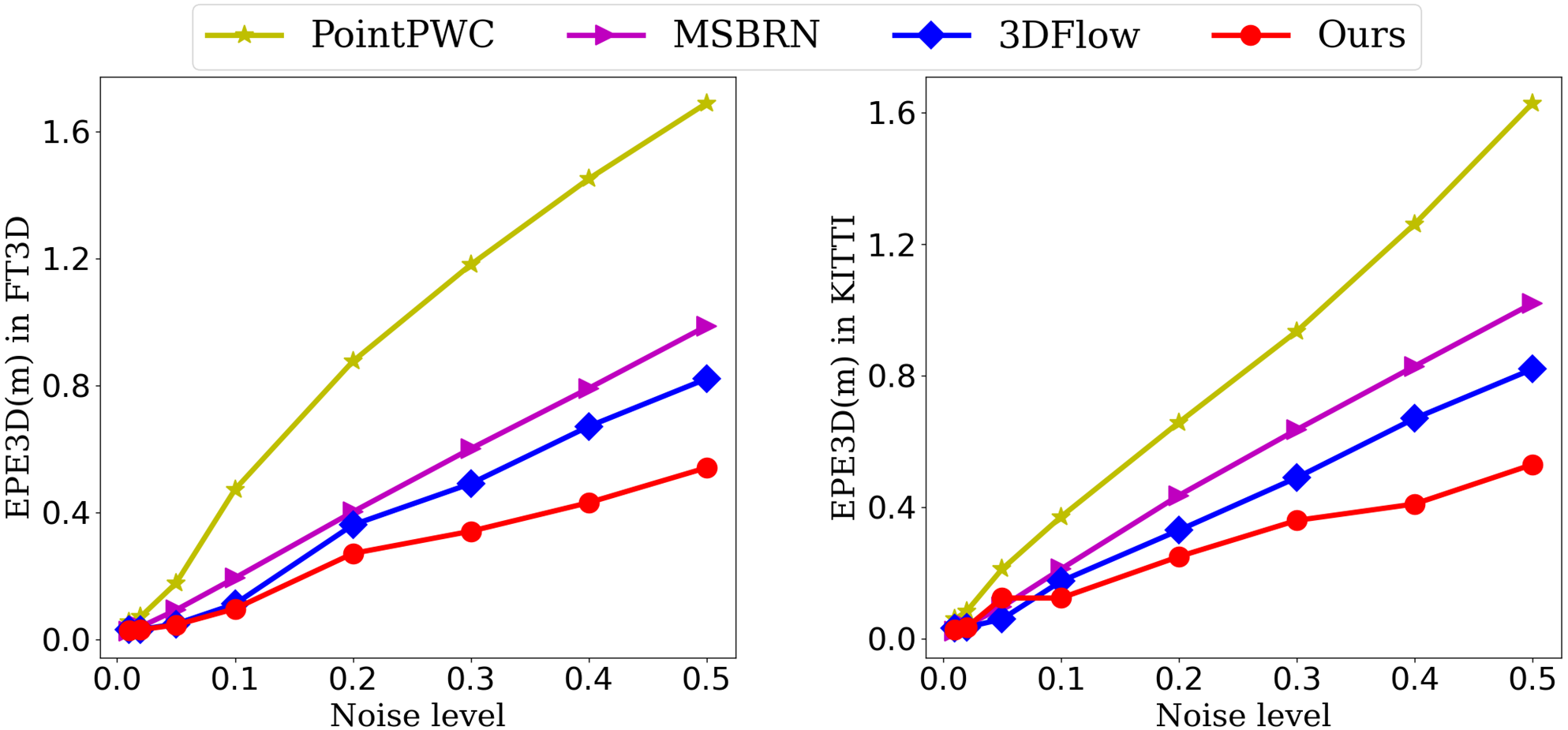}
\vspace{-5mm}   
\caption{\textbf{Adding random Gaussian noise on input points.} We add random Gaussian noise on input points, where the horizontal coordinate represents the standard deviation of the noise.}
\vspace{-4mm}
\label{fig:noise}
\end{figure}

\begin{figure}[t]
\centering
\includegraphics[width=1.0\linewidth]{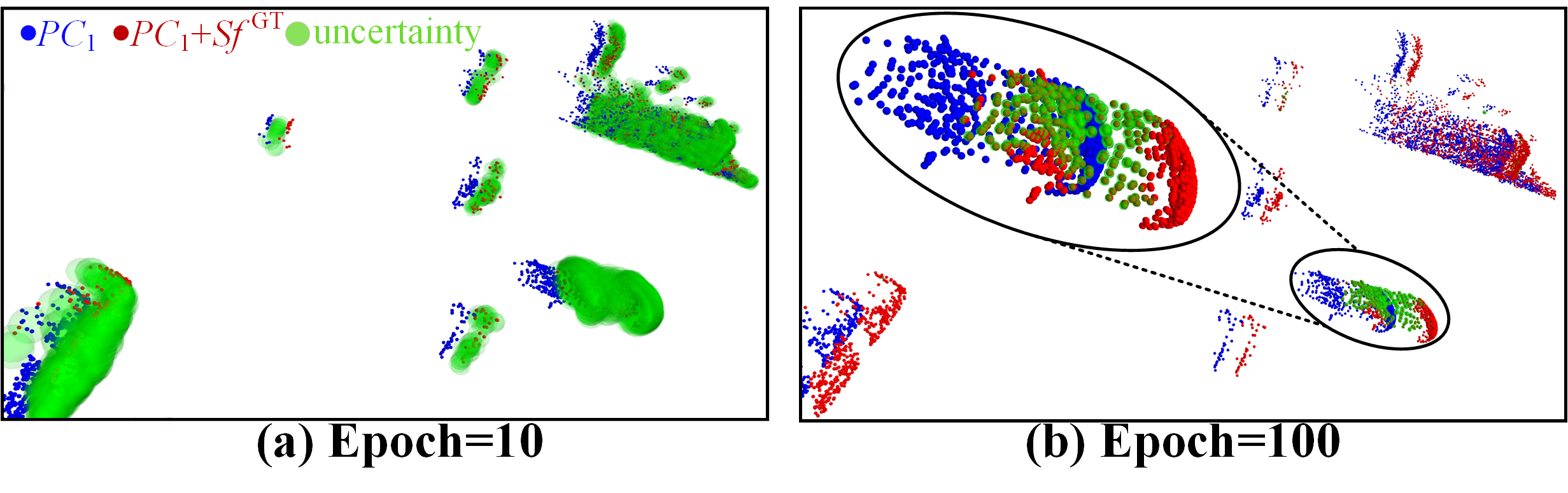}
\vspace{-7mm}
\caption{\textbf{How uncertainty works during the training process.} We visualize the uncertainty intervals respectively at different training stages (epoch=10 and epoch=100).}
\vspace{-4mm}
\label{fig:work}
\end{figure}

\section{Discussion}
\vspace{-2pt}
We discuss the motivations and reasons why our proposed diffusion-based refinement module works.

\textbf{Why diffusion?} As illustrated in Fig. \ref{fig:vis}, the introduction of diffusion model can improve the robustness for some challenging cases. We further show two quantitative comparisons by adding random noise in Fig. \ref{fig:noise}. Previous works all witness a dramatically increasing EPE3D, but our method has a relatively small fluctuation in accuracy.

\textbf{How uncertainty works?} Our uncertainty represents the reliability and confidence level of per-point predicted scene flow, which can progressively constrain the flow residual generation range. As in Fig. \ref{fig:work}, there are large uncertainty intervals for all points at the initial stages of the training. After training for 100 epochs as in Fig. \ref{fig:work} (b), all uncertainty intervals are narrow obviously because our network has better estimation ability. Furthermore, uncertainty intervals also vary in terms of different objects. As shown in Fig. \ref{fig:work} (b), there are relatively larger uncertainty intervals for the dynamic car, which are relatively difficult to learn due to its inconsistent motion. But other static objects have almost no uncertainty estimations.


\section{Conclusion}
\vspace{-2pt}
In this paper, we innovatively propose a diffusion-based scene flow refinement network with uncertainty awareness. Multi-scale diffusion refinement is adopted to generate fine-grained dense flow residuals. To improve the robustness of our estimation, we also introduce a per-point uncertainty jointly generated with scene flow. Extensive experiments demonstrate the superiority and generalization ability of our DifFlow3D. Notably, our diffusion-based refinement can serve as a plug-and-play module into previous works and shed new light on future works.

{\bf \small  Acknowledgement.} {\small This work was supported in part by the Natural Science Foundation of China under Grant 62225309, 62073222, U21A20480, and 62361166632. }


{
    \small
    \bibliographystyle{ieeenat_fullname}
    \bibliography{main}
}


\end{document}